\documentclass[runningheads]{llncs}
\usepackage{graphicx}
\usepackage{amsmath,amssymb} 
\usepackage{color}
\usepackage{booktabs}
\usepackage{float}
\usepackage{url}
\makeatletter
\g@addto@macro{\UrlBreaks}{\UrlOrds}
\makeatother
\usepackage{footnote}

\makesavenoteenv{tabular}
\makesavenoteenv{table}

\def\onedot{.}

\def\etal{\emph{et al}\onedot}

\begin{document}
\pagestyle{headings}
\mainmatter

\def\ACCV20SubNumber{4010}  

\title{A Weakly Supervised Convolutional Network for Change Segmentation and Classification} 
\titlerunning{W-CDNet}
%
\author{Philipp Andermatt\orcidID{0000-0003-3503-8693} \and
Radu Timofte\orcidID{0000-0002-1478-0402}}
\authorrunning{P. Andermatt and R. Timofte}
%
\institute{Computer Vision Lab, ETH Zürich, Switzerland \\
\email{\{anphilip,timofter\}@ethz.ch}}

\maketitle

\begin{abstract}
Fully supervised change detection methods require difficult to procure pixel-level labels, while weakly supervised approaches can be trained with image-level labels. However, most of these approaches require a combination of changed and unchanged image pairs for training. Thus, these methods can not directly be used for datasets where only changed image pairs are available. We present W-CDNet, a novel weakly supervised change detection network that can be trained with image-level semantic labels. Additionally, W-CDNet can be trained with two different types of datasets, either containing changed image pairs only or a mixture of changed and unchanged image pairs. Since we use image-level semantic labels for training, we simultaneously create a change mask and label the changed object for single-label images. W-CDNet employs a W-shaped siamese U-net to extract feature maps from an image pair which then get compared in order to create a raw change mask. The core part of our model, the Change Segmentation and Classification (CSC) module, learns an accurate change mask at a hidden layer by using a custom Remapping Block and then segmenting the current input image with the change mask. The segmented image is used to predict the image-level semantic label. The correct label can only be predicted if the change mask actually marks relevant change. This forces the model to learn an accurate change mask. We demonstrate the segmentation and classification performance of our approach and achieve top results on AICD and HRSCD, two public aerial imaging change detection datasets as well as on a Food Waste change detection dataset. Our code is available at: \url{https://github.com/PhiAbs/W-CDNet}
\end{abstract}

\section{Introduction}
\label{sec:introduction}
Change detection~\cite{ChangeDetectionStateOfTheArt,ChangeDetectionTechniques} is an important computer vision task. It has applications in remote sensing~\cite{ChangeDetInHeterogeneousImages,DeepFeatureRepresentation}, video surveillance~\cite{CDforSurveillance} and street view imaging~\cite{StreetChangeDetectionSakurada,StreetViewChangeDetection}, amongst others. It is a challenging task since one has to distinguish relevant changes between two temporally different images from noise as well as from irrelevant semantic changes. An image pair can belong to one of two classes: \textit{Changed} which means that there are relevant changes between the two images, or \textit{unchanged} which means that there are either no changes at all or only irrelevant ones. The user has to define in advance what counts as relevant change.

Many data-driven change detection methods are trained with pixel-level labels~\cite{StreetChangeDetectionSakurada,StreetViewChangeDetection,FCN_Siamese}. However, creating a pixel-level change detection dataset is costly and time-consuming since two images have to be compared by hand and labeled on pixel-level. As a result, there is a need for change detection approaches which can be trained with simpler labels, for example image-level labels or bounding boxes. Additionally, many weakly supervised change detection approaches need both changed and unchanged images for training. This reduces their usability to datasets where unchanged images are actually available or can be generated with image augmentation.

\begin{figure}[tbh!]
    \centering
    \includegraphics[width=1.0\textwidth]{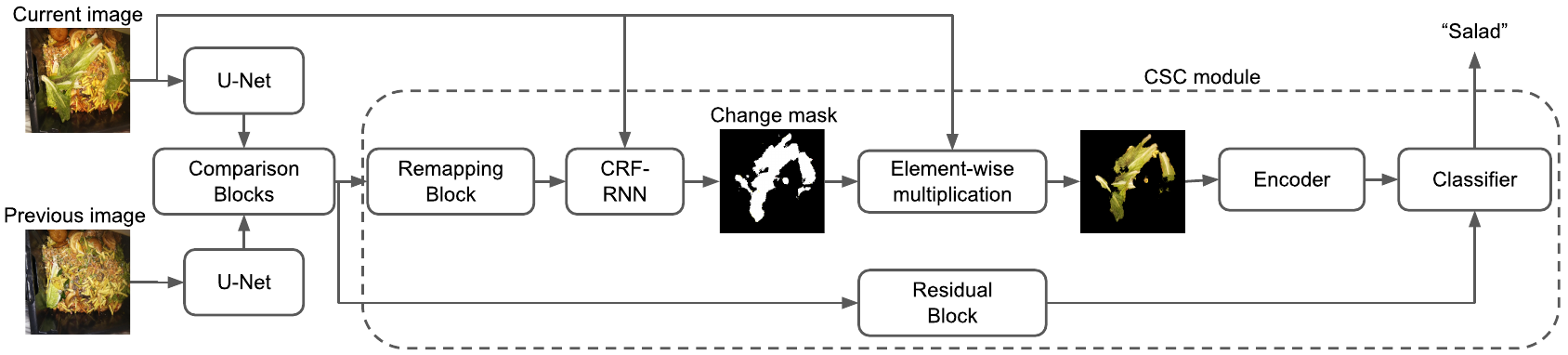}
    \caption{Overview of the proposed W-CDNet solution.}
    \label{fig:model_very_simple_intro}
\end{figure}

In this work, we propose W-CDNet, a data-driven change detection network that can be trained with image-level semantic labels. By image-level semantic labels we mean that the labels describe the changed object and not only the binary state of the image pair (\textit{changed} or \textit{unchanged}, as used in other works~\cite{Khan2017LearningDS,DenseHighResSiameseNetwork,SimpleBackgroundSubtractionConstraint}). Thus, the label \textit{changed} is further split up into several semantic labels like \textit{apple} or \textit{bread} in case of a Food Waste dataset. 

In contrast to other approaches~\cite{Khan2017LearningDS,DenseHighResSiameseNetwork,SimpleBackgroundSubtractionConstraint}, our method can be trained in one of two ways: (i) With changed image pairs only, meaning that every image pair in the dataset contains at least some relevant change; (ii) With a mix of \textit{changed} and \textit{unchanged} image pairs.

Fig.~\ref{fig:model_very_simple_intro} shows a high-level overview of our W-CDNet model. We use a W-shaped siamese network~\cite{SiameseNetworks} based on two U-Nets~\cite{U_Net} with shared weights for feature extraction. A group of custom Comparison Blocks compares the feature maps from the siamese network and creates a high-resolution feature map containing information about the differences between the two images. Our custom Change Segmentation and Classification (CSC) module remaps these features to create a meaningful change mask which is then used to segment the current input image. The segmented image gets encoded and, together with the feature vector from the Residual Block, gets fed to a classifier which predicts an image-level label. 

If the change mask created within the CSC module does not contain relevant change, the classifier is not able to predict the correct image-level label. At the same time, the model can not simply focus on the complete image thanks to our remapping function (part of the CSC module). Thus, the model is forced to learn a change mask which marks relevant changes and suppresses irrelevant changes, with weak supervision. Moreover, we integrate a CRF-RNN layer~\cite{crfasrnn_iccv2015} to refine the change mask created within the CSC module.
Our approach works for single-label images. Since it is trained with image-level semantic labels, we can directly classify the changed object.

The key contributions of our work include: 
\begin{itemize}
    \item[(i)] W-CDNet, a novel change detection network which can be trained with weak supervision using image-level semantic labels. To the best of our knowledge we are the first ones to use image-level \textit{semantic} labels instead of binary labels for weak supervision. 
    \item[(ii)] CSC, a novel Change Segmentation and Classification module that creates a high-resolution change mask at a hidden layer and encodes the segmented current input image. The CSC module is crucial for W-CDNet's performance.
    \item[(iii)] We show that our W-CDNet improves the state-of-the-art for weakly supervised change detection on the AICD dataset~\cite{AICDDatasetPaper}. Additionally, we are the first ones to report weakly supervised change detection results on the HRSCD~\cite{HighResSemanticChangeDetectionDataset} dataset. We also report performance on the Food Waste dataset.
    \item[(iv)] We make newly collected image-level semantic labels for the AICD dataset publicly available.
\end{itemize}

\section{Related Work}
\label{sec:relatedWork}
Change detection plays an important role in remote sensing~\cite{ChangeDetectionTechniques,ActiveLearningAndMRF,ConditionalAdversarialNetworks,HighResSemanticChangeDetectionDataset} and in street view imagery~\cite{StreetChangeDetectionSakurada,LearningToMeasureChange,StreetViewChangeDetection}. Today, with the advent of deep learning~\cite{gradientBasedDL,DeepLearningLeCun}, data-driven approaches are used most often as they lead to better performance than previous approaches~\cite{Muchoney1994ChangeDF,Lambin1996ChangeDA,osti_6621738}. These related works can be categorized into fully supervised, weakly supervised and unsupervised change detection data-driven methods.

\subsection{Fully Supervised Change Detection}
Many methods are trained in full supervision with pixel-level labels to obtain a good segmentation performance. Depending on the exact task, either binary change masks or semantic change masks are used. 

Guo~\etal~\cite{LearningToMeasureChange} proposed a siamese fully convolutional network. They subtracted the features from three hidden layers, used the result to create a change mask and proposed a new loss to punish noisy changes. Jiang~\etal~\cite{PGA_SiamNet} introduced a pyramid feature-based attention-guided siamese network using a global co-attention mechanism to emphasize correlations between input feature pairs. Daudt~\etal~\cite{FCN_Siamese} presented three fully convolutional networks for change detection. Two use a siamese architecture with shared weights as an encoder. A decoder uses features from different hidden layers from the encoder to create a change mask. Lebedev~\etal~\cite{ConditionalAdversarialNetworks} created a conditional adversarial network for fully supervised change detection. 

Bu~\etal~\cite{MaskCDNet} proposed a method for change detection in unregistered images. A first module roughly predicts change areas and matching information for an image pair and a second one refines the change areas. 
Daudt~\etal~\cite{HighResSemanticChangeDetectionDataset} proposed four methods to perform semantic change detection, all of which use the same encoder-decoder structure with residual connections. They either directly predict a semantic change mask or an individual change mask and semantic labels which are then combined in a second step.

\subsection{Weakly Supervised and Unsupervised Change Detection}
Sakurada~\etal~\cite{WeaklySupervisedSilhouetteBased} proposed a siamese network which first creates a change mask and then performs semantic segmentation of the changed regions. Both model parts are trained with pixel-level labels. They use separate change detection and semantic segmentation datasets for training. Jiang~\etal~\cite{DenseHighResSiameseNetwork} used a siamese network in combination with a conditional random field (CRF)~\cite{conditionalRandomFields}. They only use image-level labels for training. Additionally, they proposed a weighted global average pooling layer (WGAP) which allows them to jointly predict pixel-level and image-level labels.

Yu~\etal~\cite{ActiveLearningAndMRF} used active learning and markov random fields to reduce the number of needed ground truth change masks. Jong~\etal~\cite{UnsupervisedChangeDetection} trained their model on a standard segmentation dataset. They then compared extracted features of two images in order to create a difference image. 

Minematsu~\etal~\cite{SimpleBackgroundSubtractionConstraint} proposed a convolutional neural network (CNN) which uses a simple pixel-wise subtraction of the two input images as an additional constraint. Their model is trained with image-level labels only. Chianucci~\etal~\cite{SpatialTransformerNetworks} used spatial transformer networks~\cite{SpatialTransformers} and a sliding window to detect changes between two images. They did not perform change segmentation but rather simple change detection with bounding boxes. Khan~\etal~\cite{Khan2017LearningDS} used a CNN in combination with a CRF in order to simultaneously perform binary classification and create a change mask. They use a CNN to predict a first change mask and a CRF to refine the change mask. They trained their model with image-level binary labels. 

In contrast to the existing weakly supervised change detection approaches, our method can be trained with image-level \textit{semantic} labels which describe the changed object. Our proposed Change Segmentation and Classification (CSC) module allows us to create a change mask and simultaneously label the changed object for single-label images. As a result, our method can be trained with datasets which contain \textit{changed} image pairs only or with datasets which contain a mix of \textit{changed} and \textit{unchanged} image pairs. Additionally, in contrast to other works~\cite{SimpleBackgroundSubtractionConstraint,DenseHighResSiameseNetwork,Khan2017LearningDS}, our model creates a change mask and segments the image containing change at a hidden layer. The resulting segmented image is then refined using a CRF-RNN~\cite{crfasrnn_iccv2015} layer and used to predict an image-level semantic label. 
Akin to~\cite{PGA_SiamNet,LearningToMeasureChange,FCN_Siamese,WeaklySupervisedSilhouetteBased,DenseHighResSiameseNetwork}, we too employ a siamese network to extract feature maps from an image pair. However, we propose a W-shaped siamese network based on two U-Nets~\cite{U_Net} and introduce a custom Comparison Block to compare these feature maps before our novel CSC module to jointly segment and classify the change.

\section{Proposed Method}
\label{sec:method}
In this section we introduce our proposed W-CDNet. We start by providing an overview of the processing pipeline and architecture, then we provide the details for each of the major components. We also explain how the Change Segmentation and Classification (CSC) module lets the model learn an accurate change mask at a hidden layer.

\subsection{W-CDNet overview}
\begin{figure}[!ht]
    \centering
    \includegraphics[width=1.0\textwidth]{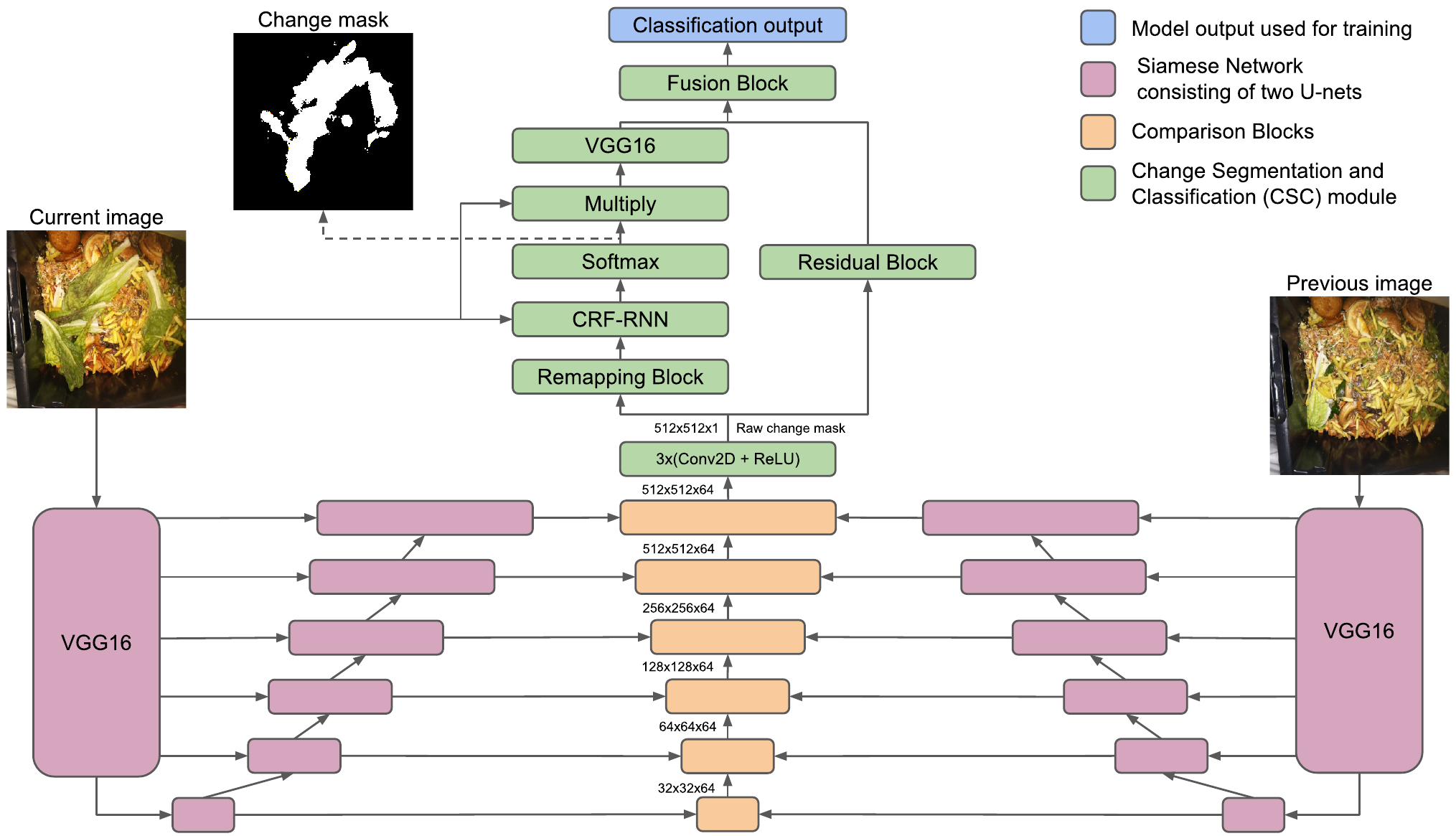}
    \caption{W-CDNet model architecture.} 
    \label{fig:model_all_simplified}
\end{figure}

We introduce W-CDNet (see Fig.~\ref{fig:model_all_simplified}), a weakly supervised convolutional network which detects and segments relevant pixel-level changes in an image pair. The proposed method was trained with image-level semantic labels, as explained in introductory section~\ref{sec:introduction}. The model takes two temporally different, co-registered RGB images as input (called \textit{previous image} and \textit{current image}). If the image pairs were not yet co-registered, we aligned them in a pre-processing step.

A siamese model (shared weights), which consists of two U-Nets~\cite{U_Net}, extracts feature maps from the image pair. These feature maps get compared by a chain of Comparison Blocks. By comparing features from several hidden layers, we leveraged both high-level and low-level information which allowed us to filter out unwanted changes while still creating an accurate high-resolution change mask. 

The output of the last Comparison Block is a raw change mask, which gets processed by the CSC module. The CSC module is the core of W-CDNet: It lets the model learn a change mask at a hidden layer. The CSC module and the individual blocks are described in more detail in the following sections.

\subsection{Siamese Network}
The Siamese Network is based on U-Net~\cite{U_Net} which uses a VGG16~\cite{VGG16} model as feature extractor. An image pair is used as model input. The U-Nets extract high-level and low-level feature maps from the two images. The feature maps generated before every up-sampling layer are passed on to the Comparison Blocks.

\subsection{Comparison Block}
This block compares feature maps from the two U-Nets. It highlight differences of relevant features between the two images. Fig.~\ref{fig:blocks_more_detailed_cropped} shows the different layers of a single Comparison Block.
We used a total of six Comparison Blocks, each of which compares features from a different U-Net layer. This allows the model to use both high-level and low-level information to create a change mask.
The feature maps created by one Comparison Block get up-sampled and are fed to the next Comparison Block. The feature maps from the last block get processed by three convolutional layers with ReLU~\cite{reLU} activation, which then output a single-channel raw change mask.

\subsection{CSC module}
The CSC module is the core of W-CDNet. It is split up into two independent branches. One branch, consisting only of a Residual Block, extracts features which help to decide whether the image pair contains any relevant change at all. 

The other branch further processes the raw change mask in the Remapping Block which assumes that the highest mask activations always correspond to relevant change, independent of whether there actually is relevant change in the image pair or not. The Remapping Block creates a change mask which gets refined by the CRF-RNN~\cite{crfasrnn_iccv2015} layer. The resulting final change mask is used to segment the current input image. The segmented image then gets encoded by the VGG16~\cite{VGG16} model. The Fusion Block concatenates the feature vectors from the two branches and predicts an image-level label. The model has to highlight relevant changes, else the Fusion Block will not be able to predict the correct image-level label. We now describe the blocks from the CSC module in more detail.


\subsubsection{Remapping Block.}
The Remapping Block receives the raw change mask from the Comparison Blocks. The change mask pixels with a high activation (relative to the other pixels) are assumed to mark relevant change. This is enforced by the following formulas:
\begin{equation}
    \hat{X}_{map} = \left( \frac{\hat{X} - min(\hat{X})}{max(\hat{X}) - min(\hat{X})} \cdot \alpha \right) - \frac{\alpha}{2}
    \label{eq:X_map}
\end{equation}

\begin{equation}
    \hat{Y} = sig \left( \hat{X}_{map} \right)
    \label{eq:sigmoid_map}
\end{equation}
where $\hat{X}$ is the raw change mask which can contain values from the interval $\left[ 0,\infty \right]$, $\alpha$ is a hyper-parameter, $sig()$ is the sigmoid function and $\hat{Y}$ is the intermediate predicted change mask which now contains values from the interval $\left[ 0, 1 \right]$.

Equation~\eqref{eq:X_map} shows the intermediate result of the remapping process. Independent of the input range, the values of $\hat{X}_{map}$ lie in the interval $\left[ - \frac{\alpha}{2}, \frac{\alpha}{2} \right]$ where $min(\hat{X})$ gets mapped to $- \frac{\alpha}{2}$ and $max(\hat{X})$ gets mapped to $\frac{\alpha}{2}$ (Exception: All features in $\hat{X}$ have the exact same value). Since these values are then fed into a sigmoid function, the hyper-parameter $\alpha$ influences how strongly the values marking relevant change get separated from the values marking irrelevant change. The smaller $\alpha$ is, the smoother the distribution of the change mask values will be. We set $\alpha$ to 32 for training (without CRF-RNN layer) and to 16 for finetuning (including CRF-RNN layer).

The intermediate change mask can now be further refined with a CRF-RNN~\cite{crfasrnn_iccv2015} layer. The final change mask then gets multiplied element-wise with the current input image, which results in an image where only regions with relevant change are visible. This step, combined with the remapping explained above, forces the model to learn an accurate change mask. The segmented image is then fed to a VGG16~\cite{VGG16} model which creates a feature vector describing the region marked by the change mask.

\subsubsection{Residual Block.}
While the Remapping Block assumes that there is always \textit{some} relevant change between the two input images, the Residual Block helps to decide if this is actually true or not.
To this end the Residual Block takes the raw change mask as input and creates a feature vector which helps the model to distinguish between changed and unchanged image pairs. Without the Residual Block, the model is not able to do this distinction due to the Remapping Block. It can, however, still learn to segment changes when trained on changed images only. Fig.~\ref{fig:blocks_more_detailed_cropped} shows the different layers of the Residual Block. 

\subsubsection{Fusion Block.}
\begin{figure}[!ht]
    \centering
    \includegraphics[width=1.0\textwidth]{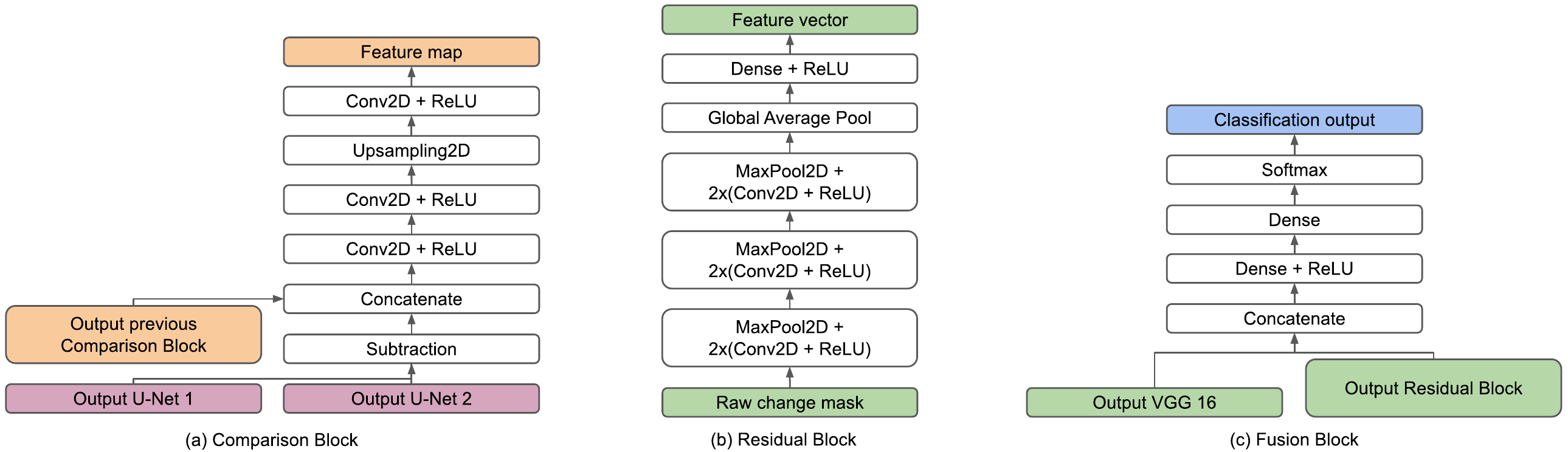}
    \caption{Comparison Block, Residual Block and Fusion Block in more detail. The colors of the block inputs and outputs refer to the ones in Fig.~\ref{fig:model_all_simplified}.}
    \label{fig:blocks_more_detailed_cropped}
\end{figure}

The Fusion Block concatenates the feature vectors created by the VGG16 model and the Residual Block and classifies them into $N$ classes, from which $N-1$ describe the changed object and one is the unchanged class. Fig.~\ref{fig:blocks_more_detailed_cropped} shows the different layers of the Fusion Block. 

\subsection{Model Variations}
\label{sec:modelVariations}

W-CDNet can be trained with a combination of changed and unchanged image pairs. By removing the Residual Block from the CSC module, the model can be trained with changed image pairs only. In this case, the Fusion Block has only one input and simply serves as a classifier. It is also possible to train our model with full supervision in a multitask learning setting~\cite{multitaskLearning}. In this case, one uses both the change mask and the image-level label as model outputs. 

\subsection{Training}
\begin{figure}[!ht]
    \centering
    \includegraphics[width=0.7\textwidth]{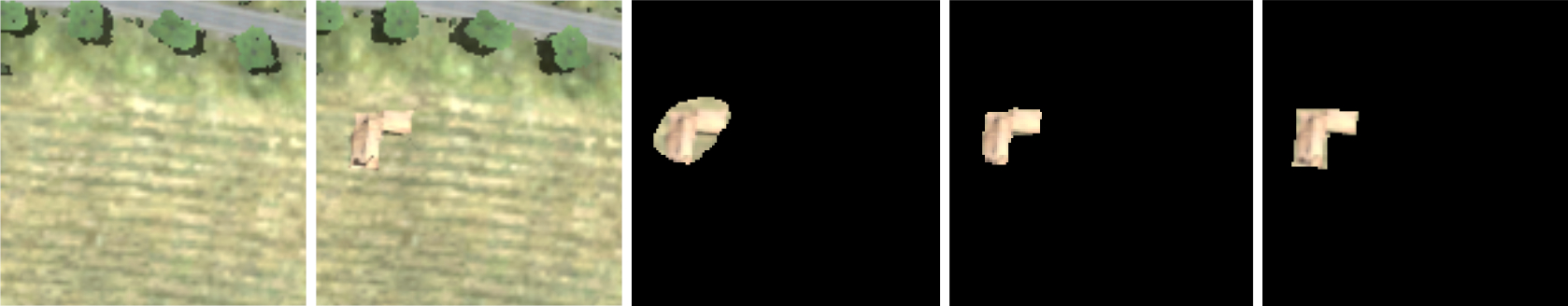}
    \caption{Influence of CRF-RNN layer on change segmentation performance, shown for an image pair from the AICD dataset. From left to right: Previous image, Current image with change object, change mask without refinement, change mask with refinement by CRF-RNN layer, ground truth change mask.}
    \label{fig:comparison_crf_no_crf}
\end{figure}

In order to speed up training, we leveraged transfer learning and used publicly available pre-trained weights for U-Net~\cite{segmentationModels}. For the VGG16\cite{VGG16} model we used publicly available pre-trained weights from a training on ImageNet~\cite{imagenet}. All other weights were initialized by sampling from a normal distribution. \\
The model has to be trained in two steps:
\\
\noindent\textbf{(1) Training:} The CRF-RNN layer~\cite{crfasrnn_iccv2015} is removed from the model. The model has to learn to correctly label the image pairs (image-level labels). During this process, it learns to create an unrefined change mask at a hidden layer. The boundaries of change objects are not accurately segmented yet, as one can see in Fig.~\ref{fig:comparison_crf_no_crf}. \\
\noindent\textbf{(2) Finetuning:} In a second step, the CRF-RNN layer is inserted into the model and the weights from step 1 for all other layers are loaded. The model is then trained again end-to-end, but with a reduced learning rate. After this training step, the changed objects get segmented more accurately thanks to the segmentation refinement by the CRF-RNN layer, as one can see in Fig.~\ref{fig:comparison_crf_no_crf}. \\

\subsubsection{Loss Function.}
We used categorical crossentropy loss to train our weakly supervised model. For fully supervised training we combined two losses: Categorical crossentropy loss for image-level semantic labels $\mathcal{L}_{IL}$ and a custom Conditional Loss $\mathcal{L}_{CM}$ for the pixel-level change mask. $\mathcal{L}_{CM}$ takes on the form of a standard binary crossentropy loss if $max(Y) > 0$, else it is set to $0$. $Y$ is the ground truth change mask. This Conditional Loss is needed since the model branch, which predicts the change mask, assumes that there is always at least some relevant change in the image pair. Thus, if the image pair does not contain any change, we do not want to induce any loss for the change mask, but only for the image-level semantic label. $\mathcal{L}_{IL}$ and $\mathcal{L}_{CM}$ are summed up to the final loss $\mathcal{L}$.

\section{Experiments and Results}
\label{sec:experimentsAndResults}
In this section we describe the experimental benchmark (datasets and performance measures), the conducted experiments, and discuss the achieved results in relation to the reported results from the literature. There is only a very limited number of weakly supervised change detection methods which report their results on publicly available datasets. Since the code for these methods is often not publicly available, we compare our results to the ones provided by the authors, by using the performance measures also used by them.

\subsection{Datasets}
\label{ssc:datasets}

Most change detection datasets contain pixel-level binary labels \cite{FCNForMultisourceBuildingExtraction,AICDDatasetPaper,CDNet2014,StreetChangeDetectionSakurada}. In our work, however, we used image-level semantic labels. We could also extract them from pixel-level semantic labels. HRSCD~\cite{HighResSemanticChangeDetectionDataset}, PSCD~\cite{WeaklySupervisedSilhouetteBased} and SCPA-WC~\cite{SemanticChangePatternAnalysis} contain pixel-level labels for semantic change detection, but only HRSCD is publicly available. We refer to \cite{ChangeDetectionStateOfTheArt} for an extended review of datasets and related literature. In this work we validated our approach on HRSCD, AICD~\cite{AICDDatasetPaper} for which we collected image-level semantic labels, and on the Food Waste dataset.

\subsubsection{Food Waste Dataset.}
The Food Waste dataset contains 57,674 image pairs of which 3,447 pairs have the image-level label \textit{unchanged}. The changed image pairs are spread over 20 classes, which results in a dataset with 21 classes, including the \textit{unchanged} class. Since the Food Waste dataset is very new, most images only contain image-level labels. A total of 145 images also contain pixel-level binary labels. They were used to evaluate segmentation performance. The image-level labels of 1,050 image pairs was double-checked by hand. These pairs were used to evaluate semantic and binary classification performance (50 of which belonged to the \textit{unchanged} class). \\
The images were resized from 3,280$\times$2,464 pixels to 512$\times$512 pixels for training and testing. The model was trained with image augmentation. 

\subsubsection{AICD Dataset.}
The \textit{Aerial Image Change Detection} (AICD) dataset~\cite{AICDDatasetPaper} is a synthetic dataset consisting of 1,000 image pairs (600$\times$800 pixels), 500 of which include hard shadows. For our experiments we only used the image pairs with hard shadows. For each image pair, there is only one object which is considered as relevant change. Since these objects are very small, we split up the images into 48 patches of size 122$\times$122 pixels with minimal overlap, which results in 24,000 image pairs. The patches were resized to 128$\times$128 pixels for training. We manually annotated the image pairs with image-level semantic labels. We will make the labels publicly available for further research.

\subsubsection{HRSCD Dataset.}
The \textit{High Resolution Semantic Change Detection} (HRSCD) dataset~\cite{HighResSemanticChangeDetectionDataset} is a very challenging dataset that contains 291 high-resolution aerial images pairs (10,000$\times$10,000 pixels). The dataset contains pixel-level semantic labels for 5 classes (Artificial surfaces, Agricultural surfaces, Forests, Wetlands, Water). Additionally, there is a ground truth change mask available for every image pair. We extracted image-level semantic labels from pixel-level semantic labels for the changed regions. 

We used 146 image pairs for training and 145 image pairs for testing. We work on image crops of size 1,000$\times$1,000 pixels which were further resized to 512$\times$512 pixels for training and testing. The original images were cropped without any overlap, which resulted in 29,100 image pairs. 

Since our model only works with single-label image pairs but HRSCD contains multi-labels, we created one label for each possible combination of the 5 semantic labels. This resulted in 31 unique semantic labels, plus the additional unchanged label. We oversampled the underrepresented classes and made sure that every class had at least 600 training samples.

\begin{figure}[htbp!]
    \centering
    \includegraphics[width=1.0\textwidth]{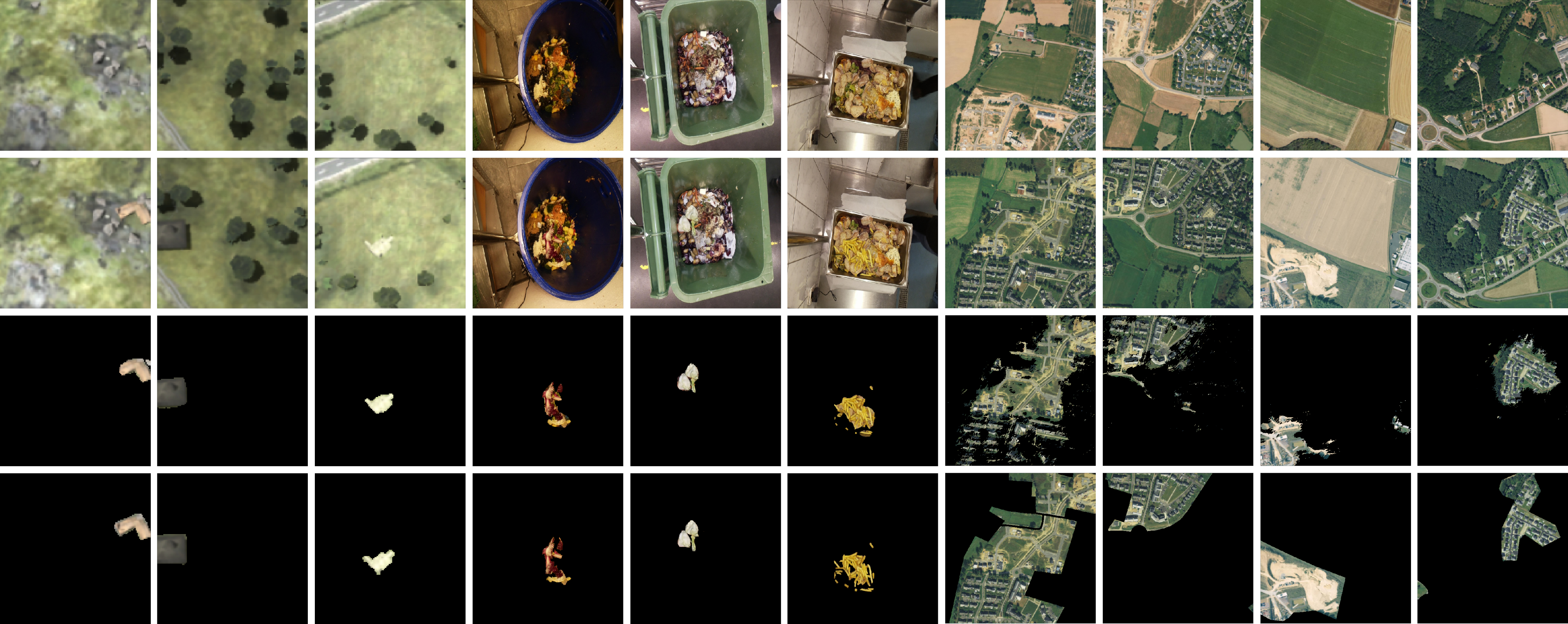}
    \caption{Qualitative results of the predicted change masks. Row 1: Previous image. Row 2: Current image. Row 3: Predicted change mask. Row 4: Ground truth change mask. Columns 1-3: AICD dataset. Columns 4-6: Food Waste dataset. Columns 7-10: HRSCD dataset.}
    \label{fig:segmentation_examples}
\end{figure}

\subsection{Performance Measures}
\label{ssc:performance_measures}
For assessing the performance we use standard measures from the literature. For classification performance we employ:
\begin{align*}
average \mkern3mu precision \mkern3mu (AP)=\sum_n (R_n - R_{n-1}) P_n,\quad accuracy=\frac{T}{N}    
\end{align*}
where $P_n$ and $R_n$ are precision and recall at the n-th threshold, $T$ is the number of all true predictions and $N$ is the total number of samples.
 
For change segmentation performance we use the standard:
\begin{align*}
    mIoU=mean(\frac{TP}{TP+FP+FN}), &\qquad Kappa=\frac{p_o - p_e}{1-p_e},\\
    Dice=\frac{2 \cdot TP}{2 \cdot TP + FP + FN},\qquad & Total \mkern3mu accuracy=\frac{TP + TN}{TP + TN + FP + FN}
\end{align*}

where $TP$ - true positive, $TN$ - true negative, $FP$ - false positive, $FN$ - false negative, $p_o$ - observed agreement between ground truth and predictions, $p_e$ - expected agreement between ground truth and predictions given class distributions.  Cohen's Kappa \cite{cohenKappa}, the S\o rensen-Dice coefficient \cite{sorensenDice} and Total accuracy are defined as in~\cite{HighResSemanticChangeDetectionDataset}.

\subsection{Influence of Residual and Comparison Blocks on Performance}
\label{ssc:residualblock_performance}

We study the importance of the Residual Block in our CSC module by training our W-CDNet with and without that Block. Since the Residual Block is needed to decide whether an image pair contains any relevant change at all, the model without this block must be trained with changed images only. Thus, we compare the change detection and segmentation performance of the full model (dataset contains also unchanged image pairs) to the model without a Residual Block (trained with changed images only). Additionally, we studied the influence on performance of different numbers of filters in the Comparison Blocks. We compare two models where the number of filters was kept constant at either 64 or 256 and a model where the number of filters decreased (256, 128, 64, 32, 16, 16 filters, from first to last Comparison Block).

We show numerical results for tests on the Food Waste dataset in Tab.~\ref{tab:numberOfFiltersInfluence}. The models without a Residual Block were trained on changed images only. The models with a Residual Block were trained on unchanged images plus the same changed images as the models without a Residual Block. We show that both models, with and without Residual Block, learn to create a change mask. They achieve similar segmentation performances. However, the influence on performance of the number of filters is very different for the two models. The model with a Residual Block shows the best segmentation results for 64 filters while the model without a Residual Block performs best with a decaying number of filters. 

Additionally, the model with a Residual Block and 256 filters did not learn to segment any changes at all. We argue that while the Residual Block is crucial for the model to distinguish between changed and unchanged images, it increases the model complexity which made training unstable for too many filters in the Comparison Blocks. 

We also show that change segmentation performance for the model without the Residual Block (and thus for a dataset without unchanged images) is on par with the performance for the model with the Residual Block. This shows that our model learns to distinguish relevant from irrelevant changes already from changed images only. 

\begin{table}[htbp!]
\centering
\caption{\label{tab:numberOfFiltersInfluence} Results on Food Waste dataset for training a model with or without the Residual Block and with different numbers of filters in the Comparison Blocks.}
\begin{tabular}{@{}lcccc|cccc@{}}
\toprule
         & \multicolumn{4}{c|}{With Residual Block} & \multicolumn{4}{c}{Without Residual Block} \\ \midrule
\multicolumn{1}{c}{\begin{tabular}[c]{@{}c@{}}Number\\ of filters\end{tabular}} &
  mIoU &
  AP &
  \begin{tabular}[c]{@{}c@{}}Top-1-\\ acc.\end{tabular} &
  \begin{tabular}[c]{@{}c@{}}Top-5-\\ acc.\end{tabular} &
  mIoU &
  AP &
  \begin{tabular}[c]{@{}c@{}}Top-1-\\ acc.\end{tabular} &
  \begin{tabular}[c]{@{}c@{}}Top-5-\\ acc.\end{tabular} \\ \midrule
64       & \textbf{46.5}     & \textbf{74.9}    & \textbf{68.2}    & \textbf{91.7}    & 45.5     & 77.7     & 72.2    & 93.0    \\
256      & 0.0      & 14.3    & 16.9    & 49.9    & 47.4     & \textbf{81.9}     & \textbf{74.4}    & \textbf{95.5}    \\
Decaying & 43.2     & 68.9    & 63.9    & 89.3    & \textbf{47.7}     & 79.4     & 72.4    & 94.6    \\ \bottomrule
\end{tabular}
\end{table}

\subsection{Influence of CRF Refinement on Performance}

In Tab.~\ref{tab:comparingCRFmethods} we compare the change segmentation performance and the binary classification performance of our W-CDNet model without any refinement to a model with a standard CRF~\cite{fullyConnectedCRFs} as post-processing step and to a model with an integrated CRF-RNN layer~\cite{crfasrnn_iccv2015} for segmentation refinement. The standard CRF was only applied after training the model without any refinement. The CRF-RNN layer was included in the training process.

We observe a significant segmentation performance increase for our model with CRF-RNN layer. The standard CRF (without parameter optimization) applied as post-processing shows only a marginal performance improvement on the AICD dataset and even significantly decreases performance on the Food Waste and HRSCD datasets. This was to be expected, as the CRF-RNN layer learns its parameters during training and adapts to each dataset individually.

\begin{table}[htbp!]
\centering
\caption{\label{tab:comparingCRFmethods} Influence on segmentation and classification performance of refining the change mask with a standard CRF or with a CRF-RNN layer.}
\begin{tabular}{@{}l|ccc|ccc|ccc@{}}
\toprule
             &         \multicolumn{3}{c|}{AICD}    &  \multicolumn{3}{c|}{Food Waste}  & \multicolumn{3}{c}{HRSCD} \\ \midrule 
             & mIoU & AP & Acc. & mIoU & AP & Acc. & mIoU & AP & Acc. \\ \midrule
No refinement  & 51.6    & \textbf{98.8} & \textbf{99.2} & 43.3 & \textbf{97.3} & \textbf{98.2} & 8.4 & \textbf{75.9} & \textbf{81.7}   \\ 
Post-processing with CRF & 52.5 & \textbf{98.8} & \textbf{99.2} & 30.3 & \textbf{97.3} & \textbf{98.2} & 4.5 & \textbf{75.9} & \textbf{81.7} \\ 
Integration of a CRF-RNN layer & \textbf{66.2} & 98.7 & \textbf{99.2} & \textbf{46.5} & 89.8 & 92.9 & \textbf{9.5} & 75.1 & 81.0 \\ \bottomrule
\end{tabular}
\end{table}

\subsection{Comparison Results on AICD Dataset}

In Tab.~\ref{tab:fullSupervisionMIOU} and Fig.~\ref{fig:segmentation_examples} we report change segmentation and binary as well as semantic change classification results on the AICD dataset~\cite{AICDDatasetPaper}. We directly compare our results to the ones reported by Khan~\etal~\cite{Khan2017LearningDS} and Chianucci~\etal~\cite{SpatialTransformerNetworks}. Additionally, we trained our W-CDNet model with full supervision in a multi-task learning setting as described in section~\ref{sec:modelVariations}. 

On AICD, we improved the state-of-the-art for weakly supervised change segmentation. However, the model from Khan~\etal~performs slightly better for full supervision. The qualitative results show that our model successfully suppresses irrelevant changes while highlighting relevant change. 

Our model had the most difficulties with objects at the edge of the images. Also, our weakly supervised model interpreted hard shadows from changed objects as part of the object itself while the ground truth masks excluded these shadows. 

\begin{table}[htbp!]
\centering
\caption{\label{tab:fullSupervisionMIOU} Change segmentation and classification results on the AICD dataset~\cite{AICDDatasetPaper}.}
\resizebox{\linewidth}{!}
{
\begin{tabular}{@{}l|ccc|cc|c|c@{}}
\toprule
                 & \multicolumn{6}{c|}{Weak supervision}                              & Full supervision \\ \cmidrule{2-8} 
 &
  \multicolumn{3}{c|}{\begin{tabular}[c]{@{}c@{}}Semantic \\ classification\end{tabular}} &
  \multicolumn{2}{c|}{\begin{tabular}[c]{@{}c@{}}Binary \\ classification\end{tabular}} &
  Segmentation &
  Segmentation \\ 
\multicolumn{1}{c|}{} &
  AP &
  \begin{tabular}[c]{@{}c@{}}Top-1\\ acc.\end{tabular} &
  \begin{tabular}[c]{@{}c@{}}Top-5\\ acc\end{tabular} &
  AP &
  Acc. &
  mIoU &
  mIoU \\ 
  \midrule
Khan~\etal~\cite{Khan2017LearningDS}     & -    & -    & -    & 97.3          & 99.1          & 64.9          & \textbf{71.0}    \\
Bu~\etal~\cite{MaskCDNet}       & -    & -    & -    & -             & -             & -             & 64.9                \\
Chianucci~\etal~\cite{SpatialTransformerNetworks} & -    & -    & -    & -             & -             & 57.0          & -                \\
\textbf{W-CDNet (ours)}     & 99.5 & 98.9 & 99.8 & \textbf{98.7} & \textbf{99.2} & \textbf{66.2} & 70.3             \\ \midrule
\end{tabular}
}
\end{table}

\subsection{Comparison Results on HRSCD Dataset}
In Tab.~\ref{tab:hrscdResultsTable} and Fig.~\ref{fig:segmentation_examples} we report change segmentation and classification results on the HRSCD dataset~\cite{HighResSemanticChangeDetectionDataset}. We are the first ones to report weakly supervised change detection performance on HRSCD. 

For full supervision we compare our results to the ones from Daudt~\etal~\cite{HighResSemanticChangeDetectionDataset}. We compare to the performance of the change detector from their \textit{strategy 3} which is trained with pixel-level binary labels. Our fully supervised model shows better performance than \textit{strategy 3} and our weakly supervised model's performance is slightly better. Daudt~\etal~reported better results than us for their \textit{strategy 4.2}, but it was additionally trained with pixel-level semantic labels and thus was stronger supervised than our method. We refer to their work~\cite{HighResSemanticChangeDetectionDataset} for the results of all strategies trained by Daudt~\etal

The change segmentation results on HRSCD are significantly lower than the ones on Food Waste or AICD. There are three main reasons for that: (i) The HRSCD ground truth change masks are generally too large: they not only contain the changed objects but they mark complete building zones; (ii) HRSCD is highly imbalanced, even more so than the AICD dataset; (iii) A part of the classes are visually very similar, which makes classification, in combination with the high class imbalance, very difficult.

\begin{table}[htbp!]
    \centering
    \caption{\label{tab:hrscdResultsTable} Change segmentation and classification results on the HRSCD dataset~\cite{HighResSemanticChangeDetectionDataset}. 
    }
    \resizebox{\linewidth}{!}
    {
\begin{tabular}{@{}l|ccc|cc|ccc|ccc@{}}
\toprule
 &
  \multicolumn{8}{c|}{Weak supervision} &
  \multicolumn{3}{c}{Full supervision} \\ \cmidrule(l){2-12} 
 &
  \multicolumn{3}{c|}{\begin{tabular}[c]{@{}c@{}}Semantic \\ classification\end{tabular}} &
  \multicolumn{2}{c|}{\begin{tabular}[c]{@{}c@{}}Binary \\ classification\end{tabular}} &
  \multicolumn{3}{c|}{Segmentation} &
  \multicolumn{3}{c}{Segmentation} \\ 
\multicolumn{1}{c|}{} &
  AP &
  \begin{tabular}[c]{@{}c@{}}Top-1\\ acc.\end{tabular} &
  \begin{tabular}[c]{@{}c@{}}Top-5\\ acc.\end{tabular} &
  AP & Acc. & Kappa & Dice &
  \begin{tabular}[c]{@{}c@{}}Total\\ acc.\end{tabular} &
  Kappa &
  \multicolumn{1}{c}{Dice} &
  \multicolumn{1}{c}{\begin{tabular}[c]{@{}c@{}}Total\\ acc.\end{tabular}} \\ 
  \midrule
\begin{tabular}[l]{@{}l@{}}Daudt~\etal~\cite{HighResSemanticChangeDetectionDataset},\\ strategy 3\end{tabular} & - & - & - & - &- & - & - & - & 12.5 & 13.8 & 94.7 \\
\textbf{W-CDNet (ours)} & 83.0 & 79.9 & 98.0 & 75.1 & 81.0 & 13.3 & 14.1 & 98.4 & \textbf{15.9} & \textbf{16.8} & \textbf{98.0} \\ \bottomrule
\end{tabular}
}
\end{table}

\section{Conclusion}
\label{sec:conclusion}
This paper presented W-CDNet, a novel data-driven convolutional network for change detection which can be trained with weak supervision using image-level semantic labels. The proposed Change Segmentation and Classification (CSC) module enables W-CDNet to learn a change mask at a hidden layer by using the custom Remapping Block and then segmenting the current input image with the predicted change mask. The segmented image is then used to predict the image-level semantic label which is used for training. In ablative studies we showed the importance of our Residual Block and of the integrated CRF-RNN layer and their impact on the W-CDNet overall performance.

We improved the state-of-the-art for weakly supervised change detection on AICD and are the first to report weakly supervised change detection results on the HRSCD dataset. On HRSCD the performance of our weakly supervised approach is on par with a related work using full supervision. 

For future work we propose to try multi-label change detection. Since our model uses image-level semantic labels for training, one could predict an individual change mask for each label. One could also try to improve information sharing within the model to improve performance. Another interesting task would be to perform change detection on unregistered images. Additionally, we hope that the fact that our model can be trained with or without unchanged image pairs opens new applications for weakly supervised change detection methods.

\bibliographystyle{splncs}
\bibliography{egbib}

\end{document}